# Hyperspace Neighbor Penetration Approach to Dynamic Programming for Model-Based Reinforcement Learning Problems with Slowly Changing Variables in A Continuous State Space

Vincent Zha[1], Ivey Chiu[1], Alexandre Guilbault[1], and Jaime Tatis[1]
[1] TELUS Communications Inc., Canada
Correspondence should be addressed to Vincent Zha

*Abstract*—Slowly changing variables in a continuous state space constitute an important category of reinforcement learning and see its application in many domains, such as modeling a climate control system where temperature, humidity, etc. change slowly over time. However, this subject is less addressed in recent studies. Classical methods with certain variants, such as Dynamic Programming with Tile Coding which discretizes the state space, fail to handle slowly changing variables because those methods cannot capture the tiny changes in each transition step, as it is computationally expensive or impossible to establish an extremely granular grid system. In this paper, we introduce a Hyperspace Neighbor Penetration (HNP) approach that solves the problem. HNP captures in each transition step the state's partial "penetration" into its neighboring hyper-tiles in the gridded hyperspace, thus does not require the transition to be inter-tile in order for the change to be captured. Therefore, HNP allows for a very coarse grid system, which makes the computation feasible. HNP assumes near linearity of the transition function in a local space, which is commonly satisfied. In summary, HNP can be orders of magnitude more efficient than classical method in handling slowly changing variables in reinforcement learning. We have made an industrial implementation of NHP with a great success.

*Keywords*—Hyperspace Neighbor Penetration, Reinforcement Learning, Slowly Changing Variable, Dynamic Programming

## I. Introduction

Slowly changing variables in a continuous state space are an important category of reinforcement learning and can be applied to many domains. For example, they can be used to model a climate control system where temperature, humidity, etc. change slowly over time. They can be also used to model chemical reaction processes where metrics vary gradually. However, this kind of problems is less addressed in recent studies.

Classical methods, such as Dynamic Programming (DP), which is the foundational approach for Model-Based Reinforcement Learning (MBRL) [1], lack the ability to model continuous state space. Variants of DP, such as function approximation, are designed to fix the problem. A variety of function approximation methods have been proposed, including simple discretization, radial basis functions, instance- and case-based approximators, and neural networks [2]. In particular, tile coding achieves remarkable success [1, 3, 4, 5, 6]. More recent explorations lead to veracious improvements such as enhanced efficiency, for example, by means of state aggregation [13], multi-grid [21], and more [7, 8, 9, 10, 11, 12, 14, 15, 22, 23, 26, 27, 28, 29]. Some in-depth studies on conditions such as convergence along with the solution exploration have been made [16, 17, 18, 19, 20, 24, 25, 30, 31].

However, such methods assume the grids granular enough so that in a transition step the state completely transits from one hyper-tile to another. These methods fail to address the environments with slowly changing variables where the variables change within a grid tile in each step.

To point out the issue, let's consider the following problem regarding a climate control system, where the continuous temperature variable is discretized with a grid system. The resolution of the grid system, i.e. the size of each tile, is set to be as large as 1 degree because of computational resource constraints. Assume the temperature is a slowly changing variable and increases by only 0.01 degree in every step. It can be seen that, if the temperature starts from 0 degree, it will stay in the same tile in the first 99 steps, and only switch into the next tile after the 100th step. The hallmark of slowly changing variable is that it takes many steps for the variable to change tile.

Facing this problem, the weakness of classical discretized DP method is apparent. By definition, in a sweep of DP policy evaluation process, each tile's value is updated with the value of another tile where the most valuable action starting from the original tile lands (together with reward, discount in consideration). However, due to the slowly changing variable, the action will land into the same tile where it starts. Therefore, the tile's value will not be affected by other tiles throughout the DP update process. Put another way, the resulting model will regard the slowly changing variable as a constant, and not be able to foresee the larger change that will be accumulated over multiple steps, which renders the model unusable.

In this paper, we introduce a Hyperspace Neighbor Penetration (HNP) approach that solves the slowly changing variable problem. The HNP is also a variant of DP based on discretized state space. However, the main advantage of HNP is that it does not require high granularity of the grid system for slowly changing variables. The main idea is that HNP captures in each step the state's partial "penetration" into its neighboring hyper-tiles in the gridded hyperspace, thus the original tile's value will be slightly affected by the neighboring tile through the DP update process, even the variable's change is much smaller than the tile size. Therefore, HNP makes the computation feasible.

HNP takes advantage of the fact that, in normal situations the transition function is near linear in a small neighborhood of the state space. This is especially true with the slowly changing variable problems, as there is unlikely a drastic change of the transition function's gradient in a small space.

## II. BACKGROUND

We start by pointing out the weakness of classical discretized Dynamic Programming (DP) methods when they deal with slowly changing variables in a continuous state space.

### A. How the Classical Discretization Method Works in Normal Environment

The term dynamic programming (DP) refers to a collection of algorithms that can be used to compute optimal policies given a fully observable model of the environment as a Markov decision process (MDP), which constitutes a Model-Based Reinforcement Learning (MBRL) problem.

That is, we assume that its state, action, and reward sets, S, A, and R, are finite, and that its dynamics are given by a set of probabilities $p(s_0, r | s, a)$, for all $s \in S, a \in A(s), r \in R$, and $s_0 \in S^+$ ($S^+$ is S plus a terminal state if the problem is episodic).

We can easily obtain the optimal policy once we have found the optimal value functions, $v_*$ or $q_*$, which satisfy the Bellman optimality equations [1]:

$$v_*(s) = \max_a E[R_{t+1} + \gamma v_*(S_{t+1}) | S_t = s, A_t = a] \quad (1)$$

where S is the set of states; $v_*$ is the optimal value of the state s; A is an action from the state; R is the reward; $\gamma$ is the discount rate.

In a deterministic environment, (1) can be simplified as

$$v_*(s) = \max_a [R_{t+1} + \gamma v_*(S_{t+1}) | S_t = s, A_t = a] \quad (2)$$

or

$$q_*(s, a) = [R_{t+1} + \gamma \max_{a'}(q_*(S_{t+1}, a')) | S_t = s, A_t = a] \quad (3)$$

where $q_*$ is the q-value of action a from state s.

As DP requires a finite amount of states, discretization is applied to the continuous variable. Thus, the state space is divided into a finite amount of tiles. In each sweep of DP, the process updates the values of all tiles. This method bears certain merits, namely it takes few hyper-parameters for tuning, involves no randomness in the calculation, and the calculation is relatively fast given a relatively small amount of tiles.

As an example, let's solve a normal problem illustrated in Fig. 1 with the classical method.

In Fig. 1, there is only one continuous variable X and the environment is a deterministic Markov Decision Process (MDP). The 1-D state space is discretized into tiles, such as $tile_0$, $tile_1$, $tile_2$, $tile_3$, and $tile_4$. The size of each tile is b, i.e. each tile covers a range of b. For example, $tile_0$ is centered at $x_0$, covering the range of $(x_0 – 0.5b, x_0 + 0.5b]$. The next $tile_1$ is centered at $x_0 + b$, covering the range of $(x_0 + 0.5, x_0 + 1.5b]$, and so on.

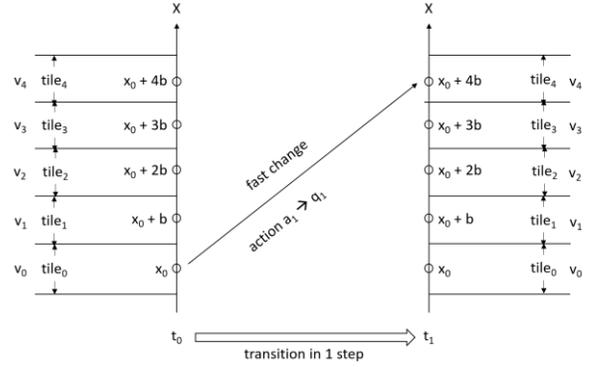

Fig. 1: Fast changing variable. Classical method works as state changes inter-tile.

The DP method iteratively updates the value of each tile in the policy evaluation process, and the policy improvement process updates the policy according to the tile values. The goal of DP is to assign the right value to each tile. Due to the discretization, all the infinite values of X within a tile's range are aggregated and treated the same in the DP process. As a result, DP can only take one value of X, normally the tile center, to represent all the infinite amount of values covered by the tile.

To clarify, we distinguish the follow 3 kinds of values.

1. The state value x, which is the value of variable X. For example, all the values of X within $tile_0$ is represented by the center value $x_0$, and $x_0$ is used as the starting value of X when calculating an action starting from $tile_0$ in the DP process.

2. The tile value v, which is calculated by DP. For example, the value of $tile_0$ is $v_0$. This is sometimes referred to as v-function value.

3. The q-value q, which is the value of an action starting from a tile. It is calculated by DP. For example, the q-value of action $a_1$ starting from $tile_0$ is $q_1$.

Fig. 1 describes a transition step from $t_0$ to $t_1$. As illustrated, X is a fast changing variable. In the transition, the state value changes from $x_0$ to $x_0 + 4b$, crossing multiple tiles. Therefore, X changes from $tile_0$ to $tile_4$. Since the transition happens inter-tile, there is no issue with the DP method. The q-value of action $a_1$ is calculated as below.

$$q_1 = r_1 + \gamma * v_4$$

where $q_1$ is the q-value of action $a_1$; $r_1$ is the reward of action $a_1$; $\gamma$ is the discount rate; $v_4$ is the value of $tile_4$.

Similarly, we can calculate the q-values of other actions starting from $tile_0$. Assuming $q_1$ is the highest among the q-values of all actions, the value of $tile_0$ will be updated with $q_1$, i.e.

$$v_0' \leftarrow q_1 = r_1 + \gamma * v_4$$

where $v_0'$ is the updated value of $tile_0$.

It can be seen that the value of $tile_4$ is "passed" to $tile_0$. This is the core mechanism of DP. By letting the values of tiles influence each other, DP is able to eventually determine the right value for each tile based on other tiles' values through iterations.

The above describes the concept of DP handling continuous variable by means of discretization. Based on that, several improvements have been made, such as state aggregation [13], multi-grid [21] to enhance efficiency, and more [7, 8, 9, 10, 11, 12, 14, 15, 22, 23, 26, 27, 28, 29], and further study on convergence [16, 17, 18, 19, 20, 24, 25, 30, 31].

### B. Weakness of the Classical Method Dealing with Slowly Changing Variables

However, the classical discretization methods directly or indirectly assume the variables in questions change large enough in a transition step. With slowly changing variables, the classical DP process will fail, as illustrated in Fig. 2.

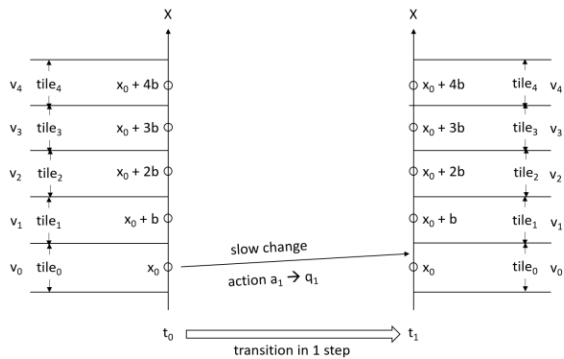

Fig. 2: Slowly changing variable. Classical method no longer works as state changes within the same tile.

Here, slowly changing variable means that the change of the variable in a transition step is too small compared to the tile size. Put another way, the grid system is not granular enough.

As illustrated in Fig. 2, by taking action $a_1$ from $tile_0$, the state value changes from $x_0$ to a value slightly greater than $x_0$ but still falls into $tile_0$. Therefore, we get

$$q_1 = r_1 + \gamma * v_0$$

$$v_0' \leftarrow q_1 = r_1 + \gamma * v_0$$

From this result we can see that, thanks to the slowly changing variable, DP uses the value of the tile to update the same tile itself, therefore DP is incapable of capturing the variable's small change in each transition, hence it prevents tile values from influencing each other. The method effectively regards the slowly changing variable as a constant, and cannot model the larger effect that will be accumulated over multiple steps, which means that the classical method fails to address the key character of slowly changing variables.

To further illustrate the issue, suppose that $tile_1$ has a very high value $v_1$, and the episode ends once the $tile_1$ is entered. Obviously, an effective model should seek entering $tile_1$, and the way to enter $tile_1$ is by entering its neighbor $tile_0$ in the first place. However, thanks to the slowly changing variable, the high value of $v_1$ cannot be passed to its neighbor $v_0$. Consequently, the model will not be able to seek moving into $tile_0$, hence not able to enter $tile_1$.

A simple idea to fix this issue might be to increase the grid's granularity, i.e. making the tiles so small that the variable changes the tile before and after the transition. However, there are two reasons preventing the idea from succeeding.

The first reason is that, the variable may change so slowly that it requires the grid system to be extremely granular. The number of tiles therefore will be too high and it makes the DP update process computationally too expensive. The second reason is more critical: in a situation where the variable's changing speed is affected by other variables or hyperparameters, it will be impossible to determine the variable's minimum change before the model is built, therefore it is impossible to determine the tile size.

### III. METHODOLOGY

#### A. Hyperspace Neighbor Penetration (HNP) Approach in a 1-D space

As such we propose the Hyperspace Neighbor Penetration (HNP) approach which is able to solve the problem with slowly changing variable, see Fig. 3.

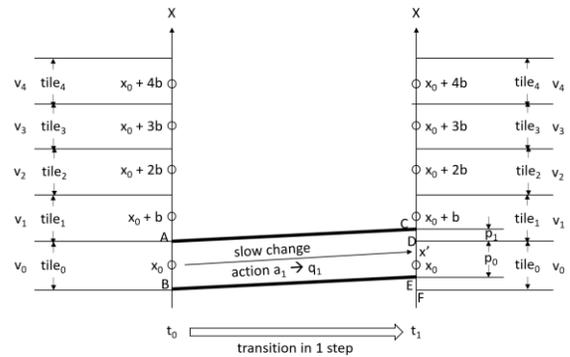

Fig. 3: Hyperspace Neighbor Penetration (HNP) approach. It can handle slowly changing variables by considering the tile's whole range of the state values.

The idea of HNP is that, unlike the classical discretization method, the state value of each tile is not represented by a single value. Instead, HNP considers the whole range of the state values covered by the tile.

Take $tile_0$ as an example, the tile covers the range (B, A] where

$$B = x_0 - 0.5b$$

$$A = x_0 + 0.5b$$

For action $a_1$ starting from B, after the transition, the resulting state value will be E, which still falls into $tile_0$, and we get

$$q_{1B} = r_{1B} + \gamma * v_0$$

where $q_{1B}$ is the q-value of action $a_1$ starting from B; $r_{1B}$ is the reward of action $a_1$ starting from B.

It can be seen that $q_{1B}$ is the same as the result of the classical method, as it only involves the value of $tile_0$.

However, for action $a_1$ starting from A, after the transition, the resulting state value will be C, which falls into the neighboring $tile_1$, and we get

$$q_{1A} = r_{1A} + \gamma * v_1$$

where $q_{1A}$ is the q-value of action $a_1$ starting from A; $r_{1A}$ is the reward of action $a_1$ starting from A.

It can be seen that the value of $tile_1$ is involved when calculating $q_{1A}$.

Next, HNP uses the 2 q-values $q_{1A}$ and $q_{1B}$ to update the tile value $v_0$ (same as before, assuming $a_1$ is better than other actions). The HNP approach assigns the weights to the 2 values as follows.

Given that the range of state values (B, A] ends up in the range (E, C] in the transition, as illustrated with the two thick lines in Fig. 3, and the range (E, D], which is denoted as $p_0$, falls into $tile_0$, whereas the range (D, C], which is denoted as $p_1$, falls into $tile_1$, the HNP assigns the weights of $p_0$ and $p_1$ to $v_0$ and $v_1$ respectively, with $p_0$ and $p_1$ being normalized as below.

$$x' = f(x_0)$$

$$p_1 = x' - x_0$$

$$p_0 = b - p_1$$

$$w_0 = p_0 / b$$

$$w_1 = p_1 / b$$

$$q_1 = w_0 * q_{1B} + w_1 * q_{1A}$$

$$v_0' = q_1$$

where x' is the state value X after the transition; f is the transition function; $p_0$ is the overlapping section between the resulting range (E, C] and $tile_0$; $p_1$ is the overlapping section between the resulting range (E, C] and $tile_1$; $w_0$ is the weight for $q_{1B}$; $w_1$ is the weight for $q_{1A}$.

HNP assumes local linearity of the transition function f among the neighboring tiles, hence $p_0$ and $p_1$ can be calculated and normalized. This assumption is commonly satisfied, as it is unlikely that the slowly changing variable change dramatically in a small local area.

It can be seen that HNP partially includes the value of the neighboring tile when updating each tile's value in the DP update process, as if the tile value "penetrates" among neighboring tiles. Therefore, HNP is able to capture the slight change of the variable without requiring the transition to be inter-tile in order for the change to be captured. As a result, HNP allows for a coarse grid system hence less amount of tiles, which makes the computation feasible.

To further illustrate the issue, suppose, as mentioned earlier, $tile_1$ has a high value, then HNP will be able to partially pass this high value to $tile_0$ and increase $tile_0$'s value, and continue to partially pass the high value to more tiles further down through the DP iterations. Therefore, the model will be able to seek moving into $tile_0$, and eventually reach $tile_1$. As a result, the model will be able to capture the tiny change of each step and foresee the larger effect accumulated over multiple steps. This way, HNP is able to handle environments with a slowly changing variable.

### B. HNP Expansion into an n-D Hyperspace

In the previous section we illustrate HNP in an environment where there is only one slowly changing variable in the state space. In this section we will illustrate how HNP can be adapted to deal with multiple slowly changing variables.

Let's see an example of an environment with two slowly changing variables as shown in Fig. 4.

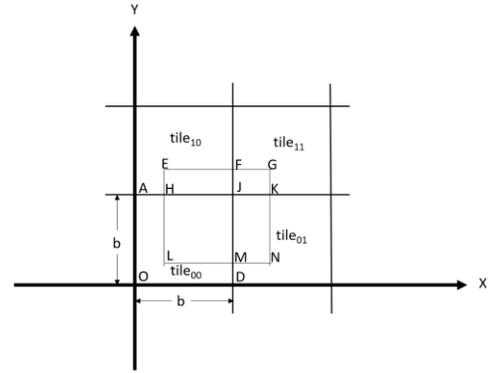

Fig. 4: HNP expansion into n-D hyperspace. A 2-D space is illustrated as an example

The two variables are named X and Y, making a 2-D continuous state space. The space is discretized into tiles by the grid system. The four neighboring tiles shown in Fig. 4 are $tile_{00}$, $tile_{01}$, $tile_{10}$, $tile_{11}$. The size of each tile is b * b, i.e. we suppose X and Y are discretized by the same unit b. It can be understood that HNP will still hold if X and Y are discretized by different units respectively.

Now let's consider the transition process for $tile_{00}$, which covers the area OAJD. Assuming after the transition, the resulting area is EGNL. It can be seen that EGNL overlaps all the four tiles. More specifically, the portion overlapping $tile_{00}$ is the area HJML; the portion overlapping $tile_{01}$ is the area JKMN; the portion overlapping $tile_{10}$ is the area EFJH; the portion overlapping $tile_{11}$ is the area FGKJ. Therefore, the q-value can be calculated as below.

$$q_{1O} = r_{1O} + \gamma * v_{00}$$

$$q_{1A} = r_{1A} + \gamma * v_{10}$$

$$q_{1J} = r_{1J} + \gamma * v_{01}$$

$$q_{1D} = r_{1D} + \gamma * v_{10}$$

$$q_1 = (HJML / b^2) * q_{1O} + (JKMN / b^2) * q_{1A} + (EFJH / b^2) * q_{1J} + (FGKJ / b^2) * q_{1D}$$

$$v_{00}' = q_1$$

where $q_{1O}, q_{1A}, q_{1J}, q_{1D}$ are the q-values of the action starting from O, A, J, D respectively; $r_{1O}, r_{1A}, r_{1J}, r_{1D}$ are the rewards of the action starting from O, A, J, D respectively; $v_{00}, v_{10}, v_{01}, v_{10}$ are the values of $tile_{00}, tile_{10}, tile_{01}, tile_{11}$ respectively; $v_{00}'$ is the updated value of $tile_{00}$.

To expand HNP into an n-D hyperspace where there are n slowly changing variables, the above method can be applied similarly. There will be $2^n$ neighboring hyper-tiles involved in the calculation, and the value of each hyper-tile will be weighed accordingly.

## IV. RESULTS

In this section we will demonstrate the power of HNP. We will first study a hypothetical problem, then refer to a successful application of HNP in a real world project.

### A. Problem Study

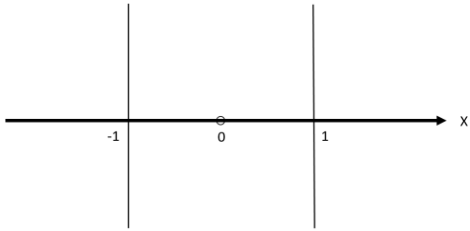

Fig. 5: Problem statement. There is only one continuous variable X. It changes slowly when going right.

We establish a hypothetical problem as shown in Fig. 5 where we can examine the HNP. In the problem there is only one continuous variable X. Below are the simple rules.

1. The agent starts from X = 0.

2. At any state, the agent can choose between two actions: either go left ($a = a_L$) or go right ($a = a_R$).

3. When $a = a_L$, the value of X will be decreased by 2, i.e. transition X ← X – 2.

4. When, $a = a_R$, the value of X will be increased by 0.02, i.e. transition X ← X + 0.02. It can be seen that X becomes a slowly changing variable when going right.

5. When X <= -1 or X >= 1, the episode finishes.

6. When X <= -1, there will be a relatively low reward of 1; when X >= 1, there will be a relatively high reward of 10.

7. There is no reward in any other situation.

We can see that the ideal policy should be keeping going right until X >= 1 to get a reward of 10 and finish the episode. Conversely, going left reaching X <= -1 and finishing the episode will be a less optimal choice, as the reward will be only 1.

### B. Classical Method Solving the Problem

Now, let us first see how the classical method of a grid system solves the problem, with the understanding that the classical method with a coarse grid system will not be able to solve the problem. For example, let us try to coarsely divide the continuous 1-D state space into only 3 tiles as follows, see Fig. 6.

- $tile_L$: (-∞, -1]
- $tile_M$: (-1, 1)
- $tile_R$: [1, ∞)

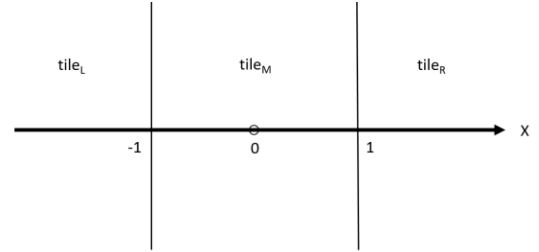

Fig. 6: Coarse grid. The continuous state space is divided into only 3 tiles.

Per Rule 6, the tile values of $tile_L$ and $tile_R$ should be

$$v_L = 1$$

$$v_R = 10$$

Neither $v_L$ nor $v_R$ changes throughout the DP iterations because the episode ends there.

Therefore, the only task of the DP is to decide the tile value of $tile_M$, i.e. $v_M$.

At the beginning, by default let us set $v_M = 0$, and then see how the $v_M$ will be updated through DP iterations.

In DP process, we first set X to the center of the $tile_M$, i.e. set X = 0. Then let us consider the two possible actions.

- For the action of going left, the transition will be X ← 0 – 2 = -2, i.e. the tile changes from $tile_M$ to $tile_L$, therefore the agent will get a reward of 1 and the episode ends.

- For the action of going right, the transition will be X ← 0 + 0.02 = 0.02. There is no tile change by this action. Therefore, the transition step goes back to the same $tile_M$.

Therefore, the problem effectively becomes Fig. 7.

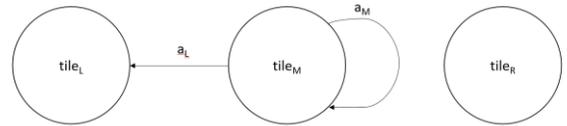

Fig. 7: Issue of the classical discretization method with a coarse grid. The value of $tile_R$ cannot affect $tile_M$. Thus, starting from $tile_M$, the agent will only go into $tile_L$.

As shown in Fig. 7, the higher value of $tile_R$ will not be able to affect the $tile_M$. When the agent starts from X = 0, the agent will only go into $tile_L$ where it receives a reward of 1 and finishes the episode (suppose there are some additional slight incentives encouraging the agent to quickly reach the end of the episode by taking action $a_L$, instead of loop back to $tile_M$ by taking action

$a_R$. For example, suppose there is a slight negative reward for each step, or there is a slight discount factor).

From the above analysis, it can be seen that the classical method with a coarse grid system fails to handle the slowly changing variable problem.

To handle such problems with classical grid method, it will require a much granular grid system. There must be at least 100 tiles in the range of (-1, 1) so that each tile covers a range of as small as 0.02. This ensures that for the action $a_R$, the X will transit into a new tile. This way, the high value of $tile_R$ will be eventually passed to the 100 tiles through the DP process, and the values of the tiles will guide the agent to always choose $a_R$ in each step, thus the agent will finally end the episode in $tile_R$ and get a higher reward of 10.

In short, the classical method will require a memory of 100 units and the DP calculation process will be also expensive.

*C. HNP Approach Solving the Problem*

By the HNP approach, we can keep the coarse grid system with only 3 tiles as illustrated in Fig. 6 and solve the problem. Let's examine how it works.

At the beginning, by default we set $v_M = 0$, then we study how $v_M$ is updated through the DP iterations

In the first iteration of DP sweep,

- For the action of going left i.e. $a = a_L$, the q-value of will be 1, because from wherever point in $tile_M$ the agent starts, the agent will enter $tile_L$. Actually this is true for any round of DP sweep iteration. Therefore

$$q_L = 1 \quad (4)$$

  where $q_L$ is the q-value of $a_L$.

- For the action of going right i.e. $a = a_R$, according to HNP, we will need to consider the whole range covered by $tile_M$, i.e. (-1, 1), instead of only consider the center point of the range as the starting point of the transition, like classical method. Therefore, we can divide the range of (-1, 1) into two sub-ranges as below.

    o When X starts from any value in the range of (-1, 0.98), after the transition the agent will end in the range of (-0.98, 1). This means X still falls in $tile_M$, and there is no tile change happened.

    o When X starts from any value in the range of [0.98, 1), after the transition the agent will end in the range of [1, 1.02). This means X has moved from $tile_M$ into $tile_R$, and will get a reward of 10.

    o According to HNP, we will need to give a weight to the reward of 10 when calculating the overall q-value of $a_R$. Since the range of [0.98, 1) only accounts for 1% of the full $tile_M$ coverage, we have

$$q_R = 99\% * v_M + 1\% * v_R$$

$$= 99\% * v_M + 1\% * 10 \quad (5)$$

where $q_R$ is the q-value of action $a_R$; $v_M$ is the tile value of $tile_M$; $v_R$ is the tile value of $tile_R$.

Note: for the consistency of writing as we emphasize the interaction of tile values, in (5) we replaced the reward of 10 with the value of $tile_R$, which is 10, as $tile_M$ is just next to $tile_R$ which is the end of the episode. In normal cases, when calculating q-value of an action, the reward and the next tile value should be calculated separately.

It can be seen from (5) that, the high value of $tile_R$, which is 10, is able to affect the $q_R$. This is the key feature of HNP.

Also, according to the definition of tile value, as well as (4), (5), we have

$$v_M' = max(q_L, q_R)$$

$$= max(1, q_R) \quad (6)$$

where $v_M'$ is the updated tile value of $tile_M$.

From (5) and (6) we can see that $v_M$ and $q_R$ will make each other increase through DP iterations, and after many rounds of DP iterations, both $v_M$ and $q_R$ will be close to 10.

Now, after the DP process has been completed, all the tiles have proper values. Subsequently, the agent can start action. It is easy to see how the agent will make correct decisions as follows.

Starting from X = 0, the agent will see that $q_L = 1 < q_R = 10$, thus will choose action $a_R$. Consequently, X = 0.02 after the action $a_R$. Then the agent will see again that $q_L = 1 < q_R = 10$, thus it will still take the action $a_R$, and will end at X = 0.04, and so on. In short, the agent will correctly always go right until X = 1 when it receives the high reward of 10 and the episode ends.

*D. Comparison between HNP and Classical Method*

It can be seen that, HNP solves the problem with only 1 tile ($tile_M$), as opposed to classical method, which requires 100 tiles. Consequently, the DP process of HNP will be also much faster than that of the classical method.

In summary, for this problem, HNP is 100 times more efficient in terms of memory usage and also extremely faster than classical method.

Further, if X changes even slower when going right, for example if X only increases by 0.002 in every right-going step, the classical method will require 1000 tiles, but HNP will still allow for only one tile. Therefore, HNP will be 1000 times more efficient than classical method. In short, we demonstrate that HNP can be orders of magnitude more efficient than classical method in dealing with slowly changing variables.

*E. HNP Application in Reality*

In reality, the transition function of an environment can be more complicated than the simple fixed amount of increment. For example, the transition function can vary depending on X, and there can be some reward in each step. In this case the HNP approach will require dividing the whole range into more tiles, so that within each tile the transition function is near linear. However, this can be often satisfied by dividing the whole range into just a few more tiles, because in reality it is very unlikely that a slowly changing variable can change dramatically in a small range. Therefore, HNP allows for a coarse tile system and is much more efficient in handling slowly changing variables than classical method.

From the above discussion, HNP can be expanded in 3 aspects. In the first aspect, HNP can be applied to fast changing variables, i.e. the transition happens inter-tile. In the second aspect, HNP can be applied to an n-D hyperspace where there are n variables. In the third aspect, it can be adapted to deal with stochastic environment. Therefore, it can be seen that HNP can handle all kinds of Model-Based Reinforcement Learning (MBRL) problems that classical discretized DP method can handle, whereas HNP possesses its unique advantage, namely the ability to deal with slowly changing variables.

We have applied HNP to an indoor climate control problem of a data center air conditioning system and effectively saved energy expenditures. In this project, the temperature is a slowly changing variable. We have established a very coarse grid system using the HNP approach. It often takes tens or hundreds of steps for the temperature to move from one tile to another. However, HNP has been able to catch the tiny change in each step and effectively modeled the long-term effects accumulated through many steps. The result has been very successful.

Our paper [32] on this successful industrial application of energy saving project based on HNP has been accepted by the Canadian AI 2021 Conference. We will be presenting the HNP implementation in the Industry Track of the Conference at the end of May, 2021.

## V. CONCLUSIONS

In this paper, we point out the weakness of classical discretized DP method when it handles environments with slowly changing variables. We then introduce the Hyperspace Neighbor Penetration (HNP) approach.

We demonstrate that HNP can be orders of magnitude more efficient than classical method in dealing with Model-Based Reinforcement Learning problems with slowly changing variables. We have made an industrial implementation of NHP with a great success.

Future work includes several aspects. For example, the current study is based on deterministic environment. HNP can be easily expanded into a stochastic environment. For another example, HNP can be expanded into continuous action space; this front will be more challenging.